\documentclass[10pt,twocolumn,letterpaper]{article}

\usepackage[pagenumbers]{wacv}

\usepackage{dblfloatfix}

\definecolor{wacvblue}{rgb}{0.21,0.49,0.74}
\usepackage{tabularx}
\usepackage{array}
\usepackage[pagebackref,breaklinks,colorlinks,allcolors=wacvblue]{hyperref}
\usepackage{booktabs}
\usepackage{multirow}
\usepackage{bm}
\usepackage{algorithm}
\usepackage{algpseudocode}
\usepackage{amsmath}
\usepackage{amssymb}
\usepackage{graphicx}
\usepackage{makecell}
\newcommand{\cmark}{\checkmark}
\newcommand{\xmark}{$\times$}
\newcommand{\pmark}{$\triangle$}

\def\wacvPaperID{634} %
\def\confName{WACV}
\def\confYear{2027}

\title{STeP-C: Real-Time Neural Compression for Distributed Fiber-Optic Sensing at Scale}

\author{Tingfeng Li \qquad Ming-Fang Huang \qquad Shaobo Han \qquad Yue Tian \\
NEC Laboratories America\\
{\tt\small \{tli, mhuang, shaobo, ytian\}@nec-labs.com}
}

\AddToShipoutPicture*{%
  \AtPageLowerLeft{%
    \put(0,28){\makebox[\paperwidth]{\parbox{0.84\paperwidth}{\centering\scriptsize
      \copyright~2027 IEEE. Personal use of this material is permitted. Permission from IEEE must be obtained for all other uses, in any current or future media, including reprinting/republishing this material for advertising or promotional purposes, creating new collective works, for resale or redistribution to servers or lists, or reuse of any copyrighted component of this work in other works. Accepted to the 2027 IEEE/CVF Winter Conference on Applications of Computer Vision (WACV).}}}%
  }%
}
\begin{document}
\maketitle
\begin{abstract}
Distributed fiber-optic sensing (DFOS) technologies such as Distributed Acoustic Sensing (DAS) generate continuous, high-resolution spatio-temporal data at rates exceeding hundreds of megabytes per second, creating severe challenges for real-time storage and transmission. While neural compression methods can achieve strong rate-distortion performance on sensing data, prior approaches either require per-segment retraining incompatible with streaming, or have only been validated offline. None have been demonstrated end-to-end on a continuously streaming DAS acquisition system at scale. We present STeP-C (Spatio-Temporal Predictive Compression), a neural compression framework that decouples spatial encoding, temporal prediction, and entropy-optimized residual coding, trained end-to-end with a rate-distortion objective motivated by conditional entropy reduction. Critically, we identify entropy coding as the dominant bottleneck for streaming deployment. We address it through GPU-side symbol computation, cached probability tables, and parallel GIL-free arithmetic coding, together with FP16 inference and batched entropy coding, enabling a full real-time pipeline built around a GPU ring buffer. We validate this pipeline on a live DAS testbed, achieving a 98.85\% file size reduction at 31.04 dB full-band PSNR and an average throughput of 535 MB/s, exceeding the 324 MB/s requirement for 1 kHz, 81 km DAS streams. We further show that STeP-C substantially outperforms conventional codecs at practical streaming bitrates, can be applied to unseen channel ranges within the same acquisition system without retraining, and largely maintains downstream detector consistency, with 96.62\% high-confidence class agreement and 100\% Jaccard similarity on detected event types.
\end{abstract}
    
\vspace{-0.5cm}
\section{Introduction}
\label{sec:intro}
\begin{figure*}[!htbp]
    \centering
    \includegraphics[width=0.99\linewidth]{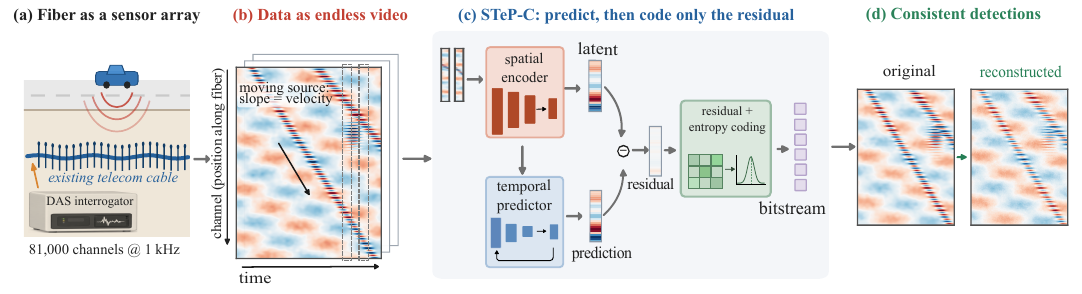}
    \vspace{-8pt}
    \caption{\textbf{STeP-C for real-time DAS compression.} \textbf{(a)} A telecom fiber acts as a dense array of sensing channels. \textbf{(b)} The DAS readout forms a spatio-temporal “video” in which physical events appear as structured patterns. \textbf{(c)} STeP-C predicts each frame and entropy-codes only the residual. \textbf{(d)} Reconstructed signals yield consistent downstream detections.}
    \label{fig:teaser}
\end{figure*}

Distributed fiber-optic sensing (DFOS) has transformed large-scale environmental \cite{lindsey2021fiber} and infrastructure \cite{ip2022distributed} monitoring by enabling dense, continuous measurements along kilometer-scale cables. In particular, Distributed Acoustic Sensing (DAS) converts an existing fiber-optic cable into an array of thousands of spatially distributed vibration-sensing channels \cite{zhan2020distributed}, including cables already carrying live commercial traffic~\cite{wellbrock2019first}, sampled at 500–1000 Hz, producing data volumes from tens of gigabytes to over a terabyte per hour. The resulting stream is naturally a two-dimensional ``waterfall" image: one axis indexes position along the fiber, the other indexes time, and each pixel is a strain-rate measurement. Practitioners commonly inspect it through this plot, because coherent physical events appear as spatially and temporally extended visual structures, most visibly diagonal streaks whose slope encodes propagation velocity, see Figure ~\ref{fig:teaser}(b). Viewed this way, a DAS interrogator acts like a camera producing a temporally unbounded video of the ground. For applications such as seismic monitoring \cite{ajo2019distributed}, cable and pipeline intrusion detection \cite{huang2021field, han2024deep}, traffic detection \cite{wellbrock2019first,li2021vehicle}, and subsea infrastructure monitoring \cite{lindsey2019illuminating}, the ability to compress and transmit or store this data in real time at acquisition is a critical operational requirement.

The visual analogy is productive but not literal. Conventional image and
video codecs are designed primarily around natural-image and video
statistics and therefore do not directly match the structure of DAS
data. JPEG 2000 primarily exploits two-dimensional spatial redundancy,
while video codecs such as H.264 additionally use block-based temporal
prediction. In contrast, DAS exhibits strongly anisotropic structure:
neighboring sensing channels form a quasi-1D spatial sequence, while
temporal variation reflects smooth signal evolution punctuated by
localized propagating events rather than apparent object motion.
Moreover, DAS compression must preserve high-dynamic-range waveform
fidelity for downstream physical inference rather than primarily
perceptual visual quality. As we demonstrate experimentally, these
mismatches limit reconstruction fidelity at the low bitrates required
for practical streaming.
What transfers from vision is therefore not the codec but the paradigm: learned nonlinear transform coding with an entropy model \cite{ball2018variational}, and predictive residual coding across frames \cite{9009574,8953892}, provided the transform, the predictor, and the entropy model are re-designed for the structure of the sensor data.

Existing DAS-specific compression methods address parts of this problem, but typically trade off compression efficiency, real-time operation, general-purpose reconstruction, or adaptability to continuous streams. These limitations leave a gap for a high-ratio neural compressor that can operate end-to-end on live DAS acquisition. In this work, we present STeP-C (Spatio-Temporal Predictive Compression), a neural compression framework for DFOS data whose overall workflow is illustrated in Figure~\ref{fig:teaser}. STeP-C achieves competitive rate-distortion performance through a modular three-stage architecture: (1) a 1D convolutional spatial encoder; (2) a recurrent temporal predictor that forecasts the next latent frame from past embeddings; and (3) an entropy-optimized residual encoder trained under a learned entropy model. 

Each stage is the sensor-domain analogue of a component in learned video compression: the 1D convolutional spatial encoder plays the role of the intra-frame analysis transform, replacing 2D convolutions with convolutions along the fiber's channel dimension; the recurrent temporal predictor replaces optical-flow motion compensation, forecasting the next latent directly rather than warping a reference frame; and the residual encoder performs residual coding under a learned entropy model. The temporal predictor is motivated by conditional entropy reduction: conditioning the current latent on available temporal context cannot increase its uncertainty, and reduces it when the current latent is statistically dependent on that context. DAS signals typically exhibit strong temporal correlation, motivating predictive coding before residual entropy coding (see Sec. \ref{sec:3_1}). 

To translate this into deployable real-time throughput, we identify entropy coding as the dominant bottleneck and redesign its implementation for streaming operation. Symbol computation is moved to the GPU, probability tables are cached across calls, and arithmetic coding is parallelized across GIL-free C++ threads, together yielding a 30× speedup over the unoptimized baseline. We validate the resulting pipeline not in simulation alone, but on a live DAS testbed with 81,000 sensing channels, corresponding to approximately 81 km of fiber at 1 m channel spacing, achieving 87.30× compression and 31.04 dB full-band PSNR at an average throughput of 535 MB/s. Beyond the live testbed, we evaluate reconstruction on fiber channels not seen during training and assess whether a downstream event detector produces consistent predictions on compressed reconstructions relative to the original data.

Our main contributions are:
\begin{itemize}
    \item \textbf{Vision-centric formulation for DFOS compression.}
    We model the DAS waterfall as a spatio-temporal visual signal and adapt
    learned transform coding to its 1D spatial layout and high-dynamic-range
    statistics, with a modular architecture separating spatial encoding,
    temporal prediction, and residual coding.

    \item \textbf{Real-time neural compression system.} We identify entropy coding as the dominant throughput bottleneck and redesign the coding path with GPU-side symbol computation, cached probability tables, parallel GIL-free arithmetic coding, FP16 inference, and batched coding. The resulting system enables, to our knowledge, the first neural DAS compressor validated end-to-end in a live acquisition pipeline directly coupled to a continuously streaming DAS interrogator.

    \item \textbf{Validation on real deployments.}
    We outperform conventional and DAS-specific codecs, demonstrate reconstruction beyond the training channel range without retraining, and show that downstream detector predictions remain largely consistent after compression.
\end{itemize}

\vspace{-0.2cm}
\section{Related Work}
\label{sec:formatting}

\begin{table}[t]
\centering
\caption{
Comparison of representative DAS data compression approaches. \cmark: demonstrated; \xmark: not demonstrated;
\pmark: partial or related demonstration.
}
\label{tab:das_comparison}
\setlength{\tabcolsep}{2.5pt}

\small
\resizebox{\columnwidth}{!}{%

\begin{tabular}{lccccc}
\toprule
Method &
\makecell{General-purpose\\reconstruction} &
\makecell{No per-segment\\training} &
\makecell{Entropy\\coding} &
\makecell{Real-time\\processing} &
\makecell{Downstream\\task validation} \\
\midrule

Dong et al.~\cite{DONG2022105181}
& \cmark
& \cmark
& \xmark
& \cmark
& \xmark \\

Wang et al.~\cite{Wang:24}
& \cmark
& \cmark
& \xmark
& \xmark
& \xmark \\

Ni et al.~\cite{2024JGRH....1..247N}
& \cmark
& \cmark
& \xmark
& \pmark
& \cmark \\

Chen et al.~\cite{10833801}
& \cmark
& \xmark
& \xmark
& \xmark
& \xmark \\

Shen et al.~\cite{10229168}
& \xmark
& \cmark
& \xmark
& \cmark
& \cmark \\

Wang et al.~\cite{wang2025ultra}
& \pmark
& \cmark
& \xmark
& \xmark
& \cmark \\

Sun et al.~\cite{sun2025ans}
& \cmark
& \cmark
& \cmark
& \cmark
& \xmark \\

DASPack~\cite{10.1093/gji/ggaf397}
& \cmark
& \cmark
& \cmark
& \cmark
& \xmark \\

Ma et al.~\cite{ma2026distributed}
& \cmark
& \cmark
& \xmark
& \pmark
& \cmark \\

Gemeinhardt et al.~\cite{gemeinhardt2026quantum}
& \cmark
& \cmark
& \xmark
& \cmark
& \pmark \\

\midrule
\textbf{STeP-C (ours)}
& \textbf{\cmark}
& \textbf{\cmark}
& \textbf{\cmark}
& \textbf{\cmark}
& \textbf{\cmark} \\

\bottomrule
\end{tabular}
}

\end{table}

STeP-C extends neural compression beyond its conventional image and video settings to a domain-specific but task-agnostic streaming problem. Rather than transferring an existing visual codec, we adapt the underlying principles of learned transform coding, temporal prediction, residual entropy coding, and rate-distortion optimization to the distinct spatial and temporal regularities of DFOS data. This formulation preserves a general-purpose reconstruction rather than optimizing the representation for a predefined downstream task, allowing the compressed signal to support storage, transmission, visualization, and future analyses. More broadly, STeP-C demonstrates that learned predictive and entropy coding can exploit structured redundancy beyond natural-image statistics, while its real-time deployment exposes practical bottlenecks, particularly entropy-coding throughput and host–device transfer overhead, that are often hidden in offline evaluation. These observations may also inform neural compression systems for other large-scale, continuously streamed sensing data. Table~\ref{tab:das_comparison} summarizes how representative DAS data compression approaches compare along the deployment-oriented criteria discussed below.

\subsection{Neural Image and Video Compression}
End-to-end learned compression has achieved state-of-the-art rate-distortion performance on natural images and video. \cite{ball2018variational} introduced the hyperprior framework using nonlinear transforms and learned entropy models. For video, \cite{8953892} proposed DVC using optical flow for motion estimation, while \cite{9009574} proposed inter-frame neural compression using residual coding in the feature space. \cite{3327546.3327736} extended this with joint autoregressive and hierarchical priors. Multiple-frame prediction was explored by \cite{Lin_2020_CVPR,cheng2019learning}. STeP-C adapts the residual latent coding paradigm from these works to the DFOS domain, replacing 2D image encoders with 1D convolutional encoders suited for fiber-optic sensor arrays. Unlike the methods above, which are typically evaluated on pre-recorded video, the proposed method explicitly addresses the throughput constraints of compressing a live sensor stream in real time.

\subsection{Neural and Task-Oriented Compression Methods for DAS}
Learned models have been applied directly to DAS compression, but under assumptions incompatible with streaming. Wang et al. \cite{Wang:24} evaluate a recurrent autoencoder on pre-recorded DAS segments, without demonstrating continuous real-time compression. Chen et al. \cite{10833801} compress 2D patches with a vision transformer, reaching ratios near 50× when combined with downsampling, but train self-supervised per segment: each new 60-second window requires on the order of 30 minutes of GPU training, or roughly 10 fine-tuning iterations with transfer from a previous window. This suits archival compression of recorded campaigns but not a live stream, where a fixed model must generalize across incoming data without retraining. Shen et al. \cite{10229168} instead map a frequency-band-energy extractor into the compressed-sensing domain, detecting and classifying events at a 70\% reduction in transmitted data without reconstructing the signal; this is fast for the task it targets but produces no reusable data for storage, transmission, or analyses not anticipated at acquisition time.

\subsection{Non-Neural Compression for DAS}
Non-neural DAS compressors provide an important deployment baseline because several already support real-time operation without model training. Dong et al.~\cite{DONG2022105181} integrate lossless compression directly into the DAS storage workflow, demonstrating compression at acquisition rate. DASPack~\cite{10.1093/gji/ggaf397} combines wavelet transforms and linear prediction with arithmetic entropy coding, providing lossless or bounded-error reconstruction while remaining suitable for real-time CPU deployment. Sun et al.~\cite{sun2025ans} similarly employ asymmetric numeral systems (ANS) entropy coding for high-ratio DAS compression. These approaches demonstrate that explicit entropy-coded bitstreams and real-time processing are achievable with non-neural pipelines, while relying on hand-designed transforms and predictors rather than learned spatio-temporal representations. STeP-C targets a different operating point by combining learned predictive coding with entropy coding, general-purpose reconstruction, and real-time operation.

\subsection{From Real-Time Inference to Live Streaming Deployment}

Real-time capability has been explored at different stages of the DAS processing pipeline, but fast inference alone does not establish that a neural compressor can operate continuously at the acquisition rate. Ni et al.~\cite{2024JGRH....1..247N}, for example, demonstrate that their lightweight reconstruction model can run on a Raspberry Pi, showing the feasibility of neural processing at the edge. Their evaluation, however, benchmarks reconstruction from subsampled DAS data rather than an end-to-end compression pipeline directly coupled to a live interrogator. Classical methods such as Dong et al.~\cite{DONG2022105181} and DASPack~\cite{10.1093/gji/ggaf397} do demonstrate real-time DAS compression, but learned approaches have largely been evaluated on pre-recorded data or through component-level runtime measurements. For continuous neural compression, the complete acquisition-to-bitstream pipeline must keep pace with incoming data while handling buffering, preprocessing, encoding, and output without interrupting the stream.
To our knowledge, STeP-C is the first neural DAS compressor validated under this end-to-end live-acquisition setting.

\section{Method}

\subsection{Spatio-Temporal Predictive Architecture}
\label{sec:3_1}

\emph{Problem formulation.} Let $X = (x_0, x_1, \ldots, x_{T-1})$ denote a sequence of $T$ DAS frames, where each frame $x_t\in \mathbb{R}^C $ represents measurements from $C$ sensing channels at time step $t$. We train the compression model end-to-end by minimizing the following rate-distortion-regularization objective:
\begin{equation}
    \mathcal{L} = \lambda_r \cdot R + \lambda_d \cdot D(x_t, \hat{x}_t)
            + \lambda_{\mathrm{tp}} \cdot \mathcal{L}_{\mathrm{tp}}
    \vspace{-3pt}
\end{equation}
where $R$ is the estimated bitrate, $D(\cdot)$ measures reconstruction distortion (MSE), and $\mathcal{L}_{\mathrm{tp}}$ is the temporal prediction loss described later in this section.

\emph{Spatial Encoder.} Each frame $x_t$ is mapped to a latent feature representation $k_t = f_a(x_t)$ using a 1D convolutional encoder. Stacked convolutions along the channel dimension capture spatial correlations among neighboring sensing channels and provide the representation used for temporal prediction and residual coding.

\emph{Temporal Predictor.} 
An LSTM temporal predictor forecasts the current spatial latent from a shared temporal context available at both the encoder and decoder. At the beginning of each sequence, no temporal prediction is available, so we set $\hat{k}_0=0$. The first spatial latent is
therefore coded directly through the residual path, yielding the reconstructed latent $\tilde{k}_0$ that initializes the temporal context. For subsequent frames, the predictor operates recursively using $\tilde{k}_0$ followed by its own previous predictions:
\begin{equation}
    \hat{k}_t = g(\tilde{k}_0,\hat{k}_1,\ldots,\hat{k}_{t-1}),
    \qquad t \ge 1.
    \vspace{-4pt}
\end{equation}

This open-loop design keeps the prediction history identical at the encoder and decoder, avoids feedback of quantization error from reconstructed latents, and allows all residuals of a sequence to be entropy-coded in a single batched call. Let \(H(\cdot)\) denote entropy. Because every $\hat{k}_t$ is a deterministic function of $\tilde{k}_0$, the predictor conditions on $\tilde{k}_0$, which cannot increase the uncertainty of the current latent: $H(k_t \mid \tilde{k}_0) \leq H(k_t)$, with strict inequality when the two are statistically dependent. Since DAS signals are strongly correlated within a short sequence, the predictor can explain a substantial part of $k_t$, and the residual encoder represents the remaining prediction error.

\emph{Temporal prediction loss.}
To explicitly supervise the temporal predictor and encourage accurate
latent forecasts, we introduce an auxiliary loss directly in latent space:
\begin{equation}
  \mathcal{L}_{\mathrm{tp}} = \mathbb{E}\!\left[\,\|\hat{k}_t - k_t\|^2\,\right]
  \label{eq:predloss}
  \vspace{-6pt}
\end{equation}
This loss penalizes prediction errors independently of the rate-distortion objective, encouraging more accurate latent forecasts and smaller residuals. 

\emph{Residual encoding and entropy coding}. For each frame, the residual $r_t = k_t - \hat{k}_t$ captures the prediction error not explained by the temporal predictor. 
For $t=0$, $\hat{k}_0=0$, and hence $r_0=k_0$; the first frame is therefore encoded without temporal prediction. For $t\geq1$, only the prediction residual is passed to the residual encoder. 
A nonlinear encoder $h_a$ maps the residual to a compact latent: $z_t = h_a(r_t)$. The latent $z_t$ is quantized to $\hat{z}_t = Q(z_t)$ via uniform quantization, approximated during training by additive uniform noise $\mathcal{U}(-0.5,\,0.5)$ as a differentiable proxy~\cite{ball2018variational}. Entropy coding uses arithmetic encoding with a learned probability model $P(\hat{z}_t)$; the estimated bitrate contribution is $R = -\log_2 P(\hat{z}_t)$.

\begin{figure}
    \centering
    \includegraphics[width=0.99\linewidth]{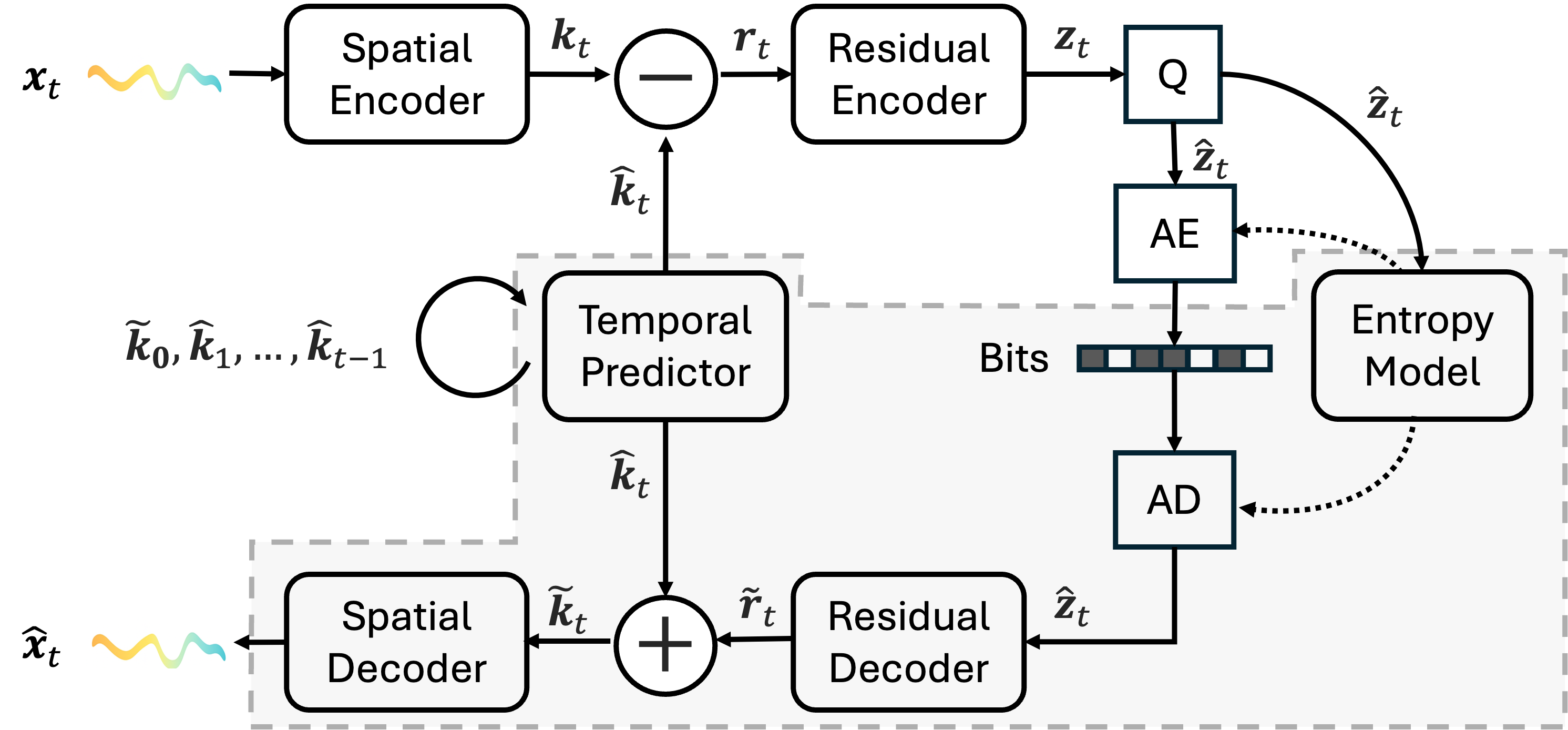}
    \caption{Spatio-temporal predictive compression architecture. The highlighted region corresponds to the components that are executed by the receiver to reconstruct the sensing data from a compressed bitstream.}
    \label{fig:stepc_arch}
\end{figure}

\emph{Decoding and Reconstruction.} At the receiver, arithmetic decoding restores $\hat{z}_t$. The residual decoder reconstructs $\tilde{r}_t = h_s(\hat{z}_t)$, which is combined with the predicted latent: $\tilde{k}_t = \hat{k}_t + \tilde{r}_t$. Finally, the spatial decoder reconstructs $\hat{x}_t = f_s(\tilde{k}_t)$. The entire pipeline is trained jointly end-to-end, as shown in Figure \ref{fig:stepc_arch}.

\subsection{Real-Time Systems Engineering}
\label{sec:method_real}
\emph{Identifying the bottleneck.} A trained STeP-C model is only deployable in real time if every stage of the pipeline sustains the target throughput. We find that the spatial encoder, temporal predictor, and residual network all run efficiently on the GPU and do not limit throughput. Instead, entropy coding becomes the primary bottleneck. Standard arithmetic coding implementations, including the one provided by CompressAI \cite{begaint2020compressai}, exhibit several inefficiencies in a streaming setting. The coder holds Python's Global Interpreter Lock during execution, so only one thread can run at a time even when calls are issued asynchronously. The probability table, expressed as a cumulative distribution function (CDF), is recomputed from float32 to int16 on every single call, even though the underlying model has not changed. Quantized symbols are computed by subtracting medians on the CPU, requiring a device-to-host transfer of the full latent tensor before the subtraction can occur. Coding also proceeds sequentially per element, with no parallelism across the batch.

\textit{An optimized entropy-coding path.} We address these inefficiencies with a custom C++ extension that reorganizes the entropy-coding path around several complementary optimizations. Symbol computation moves to the GPU: medians are subtracted from quantized values directly on-device, before any data is transferred to the CPU, eliminating a redundant round trip across the PCIe bus. The CDF table is cached: the conversion from float32 to int16 happens once, at model load or update time, and every subsequent call reuses it. The extension releases Python's GIL and performs arithmetic coding across multiple GIL-free C++ threads (8 in our deployment), enabling true parallelism across the batch. Together, these changes yield a 30× improvement in encoding throughput over the unoptimized pipeline. We apply two further accelerations on top of this fix: running the spatial encoder, temporal predictor, and residual MLP in float16 under mixed-precision autocast to reduce neural inference cost on supported hardware, and batching entropy coding across all time steps in a sequence, issuing a single coder call per batch rather than one call per time step. Section \ref{sec:thro_put} reports the cumulative throughput gains from these optimizations.

\begin{figure}
    \centering
    \includegraphics[width=0.99\linewidth]{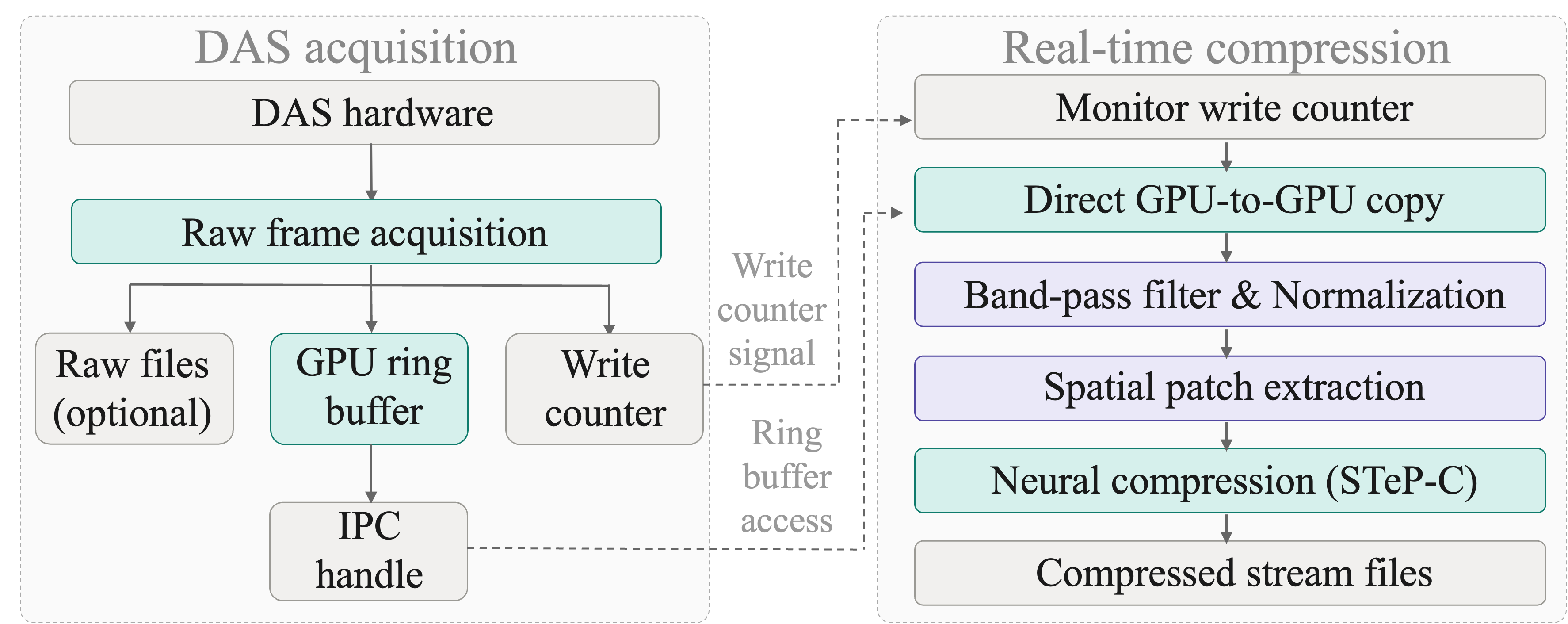}
    \caption{Real-time Compression Pipeline.}
    \label{fig:gpu_pipe}
\end{figure}

\emph{Pipeline architecture}\label{sec:systems}. Sustaining real-time throughput on a continuously arriving sensor stream requires an acquisition-aware streaming pipeline, as illustrated in Figure~\ref{fig:gpu_pipe}. We implement two concurrent processes that communicate through GPU shared memory. The acquisition process interfaces directly with the DAS interrogator, receiving raw phase frames at the full acquisition rate across all channels, and performs three tasks concurrently: writing raw frames to disk for archival (optional), writing each row into a GPU ring buffer, and incrementing a write counter in shared memory. The ring buffer holds 8,192 slots, providing approximately 8 seconds of buffering headroom, and is shared with the compression process via a CUDA interprocess communication (IPC) handle, avoiding CPU staging of the data. The compression process monitors the write counter with a lightweight polling thread, consumes new rows directly from the ring buffer via a GPU-to-GPU copy, and applies, in sequence, a streaming band-pass filter, normalization, spatial patch extraction, neural compression, and parallel entropy coding, before writing compressed output. Two of these steps require modification from their offline counterparts to remain correct under streaming. The band-pass filter is stateful, so applying it independently to each incoming chunk introduces edge artifacts at chunk boundaries; we carry the filter state across chunks by prepending the final rows of each chunk to the next as warm-up context. Normalization uses a hyperbolic tangent transform with a fixed 99th-percentile scale established offline during training, so that scaling remains consistent across chunks rather than depending on the statistics of each chunk in isolation.

\vspace{-6pt}
\section{Experiments}
\subsection{Datasets, Metrics, and Implementation}
\emph{Datasets.}
We evaluate on three real-world DAS datasets collected using our system.
The \textit{NJ Coastal Fiber Array (NJCFA)} consists of 40 recordings from
a fiber-optic cable deployment along the New Jersey coast, each containing
3,584 sensing channels. We segment the recordings into sequences of $T$
consecutive frames, with each frame spanning 512 channels, yielding 22,281
training samples and 5,565 validation samples, together with a held-out test
set of 27,916 samples. NJCFA is used for rate-distortion evaluation, comparison with
conventional and DAS-specific codecs, and architecture ablations.
The \textit{Dallas Ambient DAS (DADAS)} dataset consists of continuous recordings
collected from an urban fiber deployment in Dallas at 800 Hz across 81,000
channels. We use a 4,000-channel sub-band (channels 6,000--10,000), yielding 73,472 training samples, 18,368 validation samples, and 89,908 test samples. DADAS is used to evaluate rate-distortion performance, qualitative reconstruction, transfer to fiber channels outside the training range, and downstream detector consistency. For both NJCFA and DADAS, training, validation, and test sets are formed from separate recordings acquired at different times; no recording is shared across splits.
The \textit{Princeton Fiber Testbed (PFT)} consists of data streamed directly
from a live DAS interrogator at 1,000 Hz across 81,000 channels. PFT is used
to evaluate the complete real-time compression pipeline under live acquisition,
including end-to-end throughput, compression ratio, and full-band reconstruction
quality.

\textit{Metrics}. We report two primary compression metrics. Bits per sample (BPS) measures the average number of compressed bits per raw data element, where each element is a 32-bit float. Lower BPS indicates more compression. Peak signal-to-noise ratio (PSNR, in dB) measures reconstruction fidelity as $10 \log_{10}{(MAX^2 / MSE)}$, where MAX is the maximum amplitude of the normalized signal. Higher PSNR indicates better reconstruction. 

\textit{Implementation}.
We train four model variants, STeP-C-tiny, small, medium, and large, with residual latent dimensions $z=32,64,128,$ and $256$, respectively. Unless otherwise noted, training and evaluation use sequences of $T=10$ consecutive DAS frames, with each frame spanning 512 sensing channels. Training uses the Adam optimizer \cite{kingma2014adam} with the rate-distortion objective described in Section~\ref{sec:3_1}, with $\lambda_r=0.001$, $\lambda_d=1$, and $\lambda_{\mathrm{tp}}=0.1$. All experiments run on an NVIDIA RTX 6000 Ada GPU (48 GB) and AMD Threadripper PRO 7965WX CPU. Preprocessing follows Section \ref{sec:method_real}. For data preprocessing, NJCFA data undergo phase unwrapping followed by a 5 Hz Chebyshev Type I high-pass filter to remove low-frequency background drift. DADAS and PFT data are band-pass filtered from 1 to 200 Hz and normalized using a tanh transform with a fixed 99th-percentile scale. For streaming deployment, the band-pass filter is applied statefully: the final rows of each chunk are prepended to the next as warm-up history, eliminating edge artifacts that arise when filtering each incoming window independently.

\vspace{-3pt}
\subsection{STeP-C Rate-Distortion Performance}
Figure \ref{fig:baseline}(a) reports the rate-distortion performance of STeP-C variants on NJCFA and DADAS. Across both datasets, increasing the residual latent dimension $z$ improves reconstruction quality at the cost of higher BPS, providing a consistent capacity-rate-distortion trade-off across model variants. On NJCFA, STeP-C-small achieves 34.51 dB at 0.436 BPS, while on DADAS the same model variant reaches 41.13 dB at 0.424 BPS. The higher PSNR observed on DADAS at comparable bitrates reflects differences in the signal statistics of the two datasets, while the overall trend remains consistent across deployments.

\begin{figure}[htbp]
    \centering
    \begin{subfigure}[b]{0.99\linewidth}
        \centering
        \includegraphics[width=\textwidth]{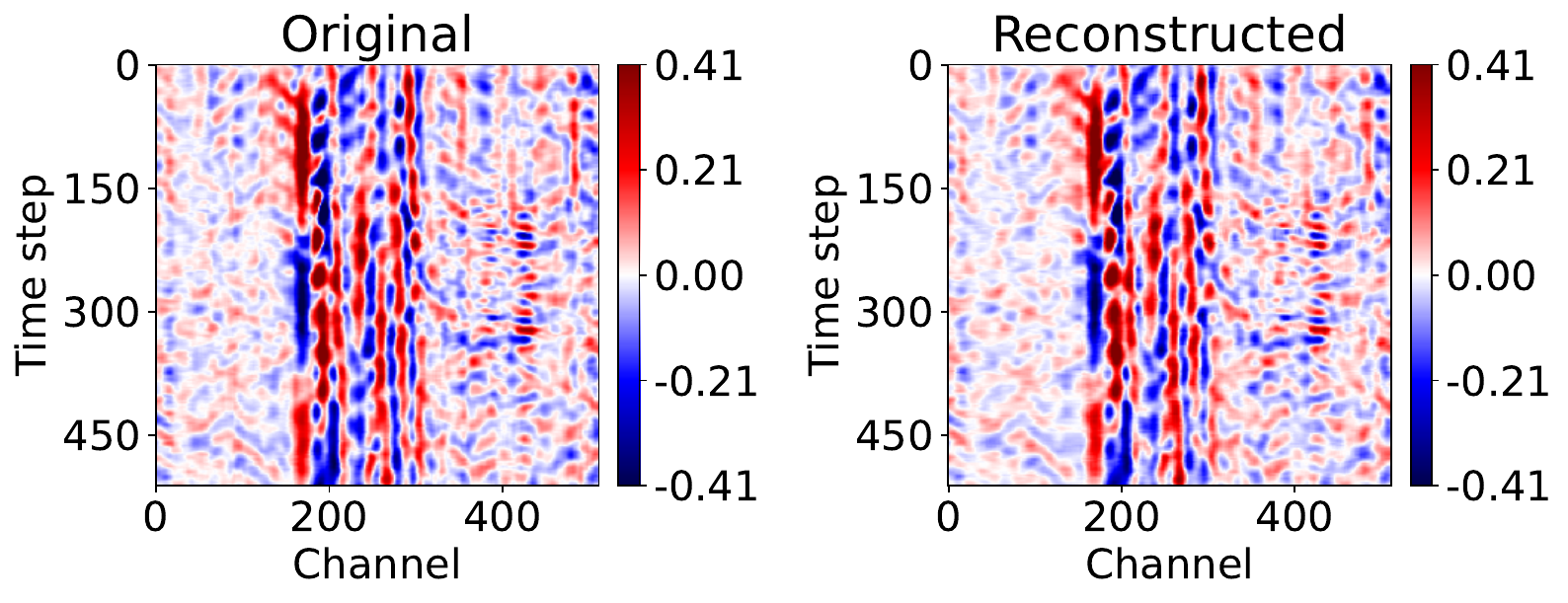}
        \caption{On trained channels, band offset 6,000 to 6,512}
        \label{fig:dadas_trained}
    \end{subfigure}

    \begin{subfigure}[b]{0.99\linewidth}
        \centering
        \includegraphics[width=\textwidth]{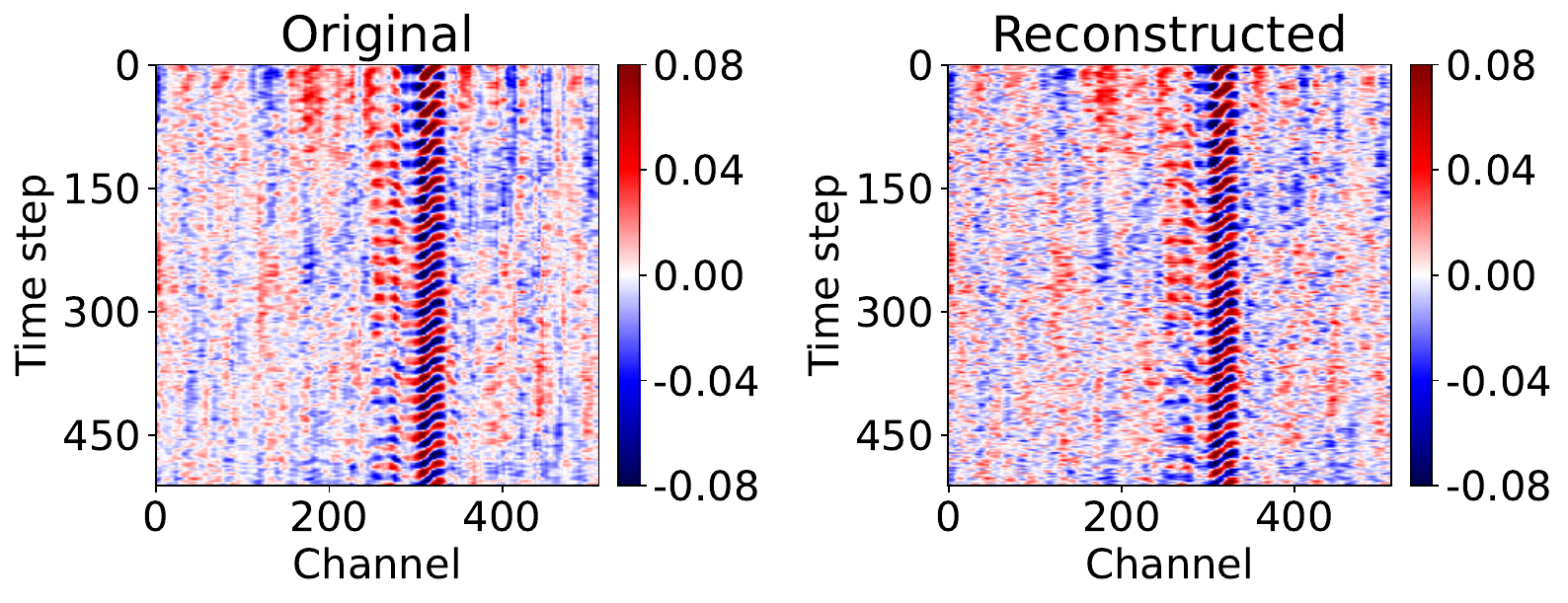}
        \caption{On unseen channels, band offset 2,000 to 2,512}
        \label{fig:dadas_new}
    \end{subfigure}
    \vspace{-6pt}
    \caption{Qualitative reconstruction results on DADAS. \textbf{(a)} Trained channel range. \textbf{(b)} Channel range excluded from training. }
    \label{fig:dadas_image}
    \vspace{-6pt}
\end{figure}

Figure \ref{fig:dadas_image} shows qualitative reconstruction results on DADAS. On trained channels (band offset 6,000 to 6,512) (\ref{fig:dadas_trained}), the reconstruction closely matches the original, preserving dominant spatial features and temporal structure. On unseen channels (band offset 2,000 to 2,512) (\ref{fig:dadas_new}), the model still recovers the primary signal structures despite not being trained on this channel range, with some smoothing of fine spatial detail. 
The full-band evaluation in Section \ref{sec:exp_live} shows a similar trend, indicating that a model trained on a localized channel range can remain effective when applied to other portions of the same acquisition system, although reconstruction quality varies with channel location.

\vspace{-0.25cm}

\subsection{Comparison with Conventional Codecs}
\begin{figure}[htbp]
    \centering
    \includegraphics[width=0.99\linewidth]{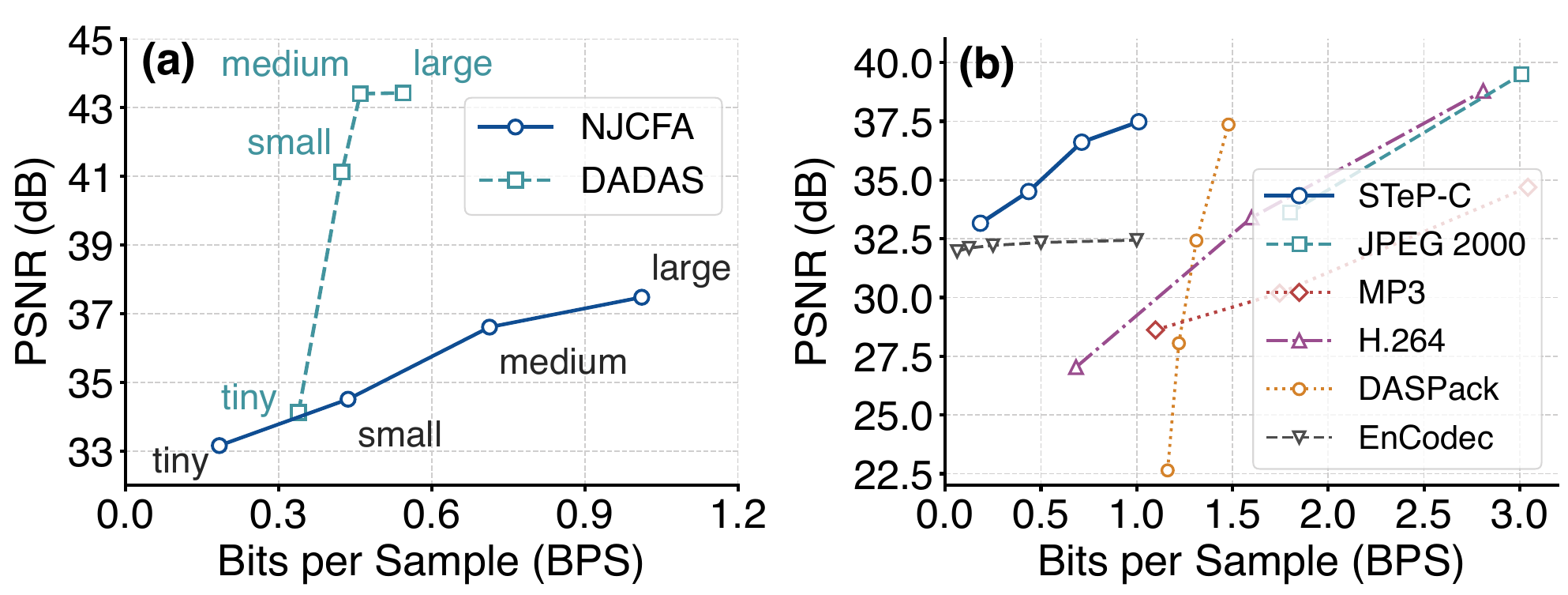}
    \vspace{-6pt}
    \caption{Rate-distortion performance across model variants and codec baselines. \textbf{(a)} Rate-distortion trade-off across STeP-C model variants on NJCFA and DADAS. \textbf{(b)} Comparison of STeP-C with conventional codecs (JPEG 2000, H.264, and MP3), a neural audio codec (EnCodec), and a DAS-specific codec (DASPack) on NJCFA.}
    \label{fig:baseline}
    \vspace{-12pt}
\end{figure}

Figure~\ref{fig:baseline}(b) compares the rate-distortion performance of STeP-C with JPEG 2000, H.264, MP3, DASPack~\cite{10.1093/gji/ggaf397}, and EnCodec~\cite{defossez2022highfi} on NJCFA. STeP-C provides operating points below 0.5 BPS while maintaining comparatively high reconstruction quality. At 0.436 BPS, STeP-C achieves 34.51 dB PSNR. EnCodec is the only other method evaluated below 0.5 BPS, but its PSNR remains near 32 dB across the tested bitrate range. DASPack operates over a relatively narrow bitrate range of approximately 1.1--1.5 BPS, with reconstruction quality increasing sharply across its tested operating points, but it does not cover the sub-0.5 BPS regime. At its lowest tested bitrate, H.264 reaches 27.03 dB at 0.68 BPS, more than 7 dB below STeP-C despite using a higher bitrate. JPEG 2000 does not produce an operating point below 1.80 BPS in our evaluation, where it achieves 33.61 dB, while STeP-C reaches 37.48 dB at 1.01 BPS. MP3 achieves 28.62 dB at 1.10 BPS and remains below STeP-C over the evaluated range.
In the practical low-bitrate regime below 0.5 BPS, STeP-C is the only evaluated method that operates in this regime while maintaining high reconstruction fidelity.
\vspace{-2pt}
\label{sec:conv_condecs}

\subsection{Architecture Ablations}
\label{sec:ablation}
\begin{figure}
    \centering
    \includegraphics[width=0.99\linewidth]{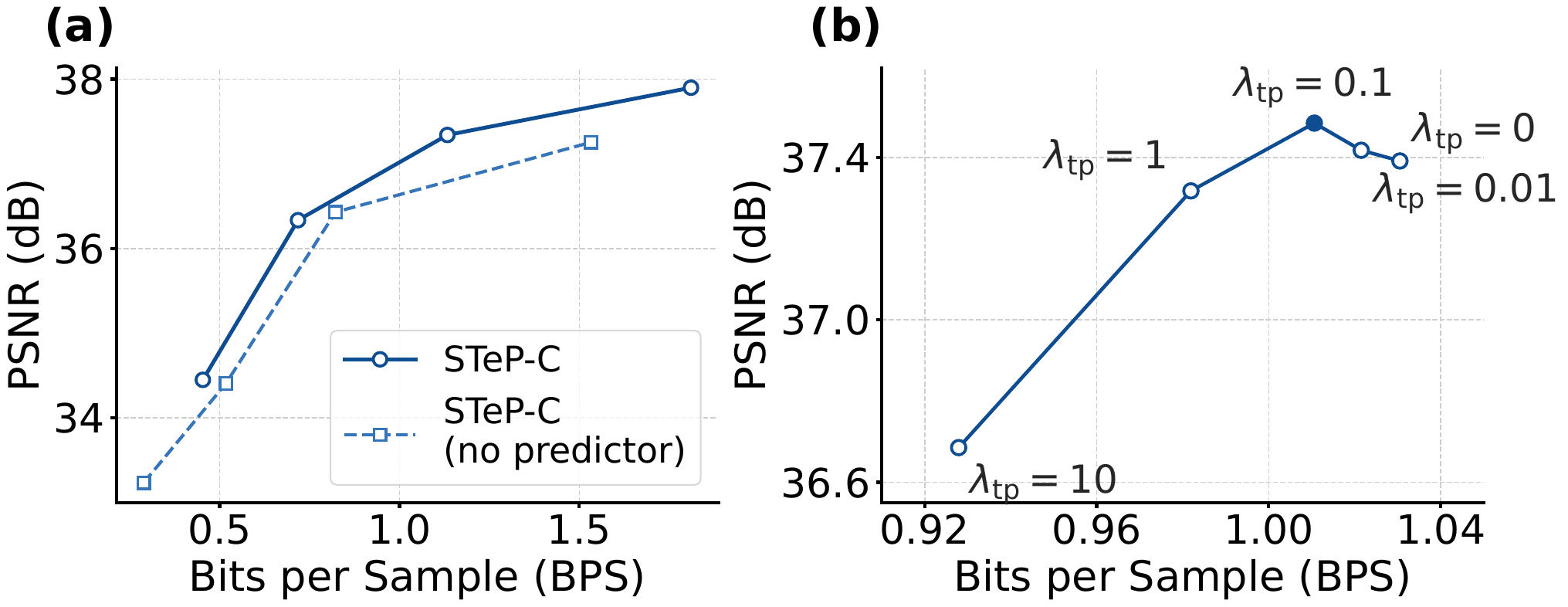}
    \vspace{-6pt}
    \caption{Architecture ablations on NJCFA.
\textbf{(a)} Rate-distortion curves of STeP-C and STeP-C (no predictor) across
four operating points, showing that the temporal
predictor shifts the curve left by up to 0.40 BPS at matched quality.
\textbf{(b)} Effect of the temporal prediction loss weight $\lambda_{\mathrm{tp}}$
on rate-distortion performance ($z=256$). The best tested value
$\lambda_{\mathrm{tp}}=0.1$ (filled circle) achieves 37.48\,dB at
1.01 BPS; excessive weighting ($\lambda_{\mathrm{tp}}=10$) reduces
bitrate at the cost of a 0.71\,dB PSNR drop.}
    \label{fig:ablation}
    \vspace{-6pt}
\end{figure}

\emph{Effect of the temporal predictor.}
Figure~\ref{fig:ablation}(a) evaluates the contribution of temporal prediction on NJCFA. At approximately 37.25 dB PSNR, the no-predictor baseline requires 1.53 BPS, while STeP-C achieves comparable quality at 1.13 BPS, corresponding to a 26\% bitrate reduction. Across the evaluated operating points, the predictor consistently shifts the rate-distortion curve
toward lower bitrates, showing that temporal prediction provides a substantial compression benefit without sacrificing reconstruction fidelity.

\emph{Effect of temporal context length.}
Table~\ref{tab:seq_len} evaluates the effect of temporal context length $T$ with all other settings fixed using the STeP-C-tiny model. Introducing temporal context with $T=3$ reduces BPS from 0.281 to 0.207 while maintaining similar PSNR (33.15 versus 33.29 dB). Increasing $T$ further to 5, 7, and 10 continues to reduce bitrate to 0.196, 0.189, and 0.184 BPS, respectively, while PSNR
remains nearly unchanged. This indicates that most of the gain comes from introducing temporal conditioning, whereas longer context provides diminishing bitrate improvements.
\begin{table}[htbp]
\centering
\caption{Effect of temporal context length $T$ on NJCFA.}
\label{tab:seq_len}
\vspace{-6pt}
\small
\begin{tabular}{lcccccc}
\toprule
$\bm{T}$ & no predictor & 3 & 5 & 7 & 10 \\
\midrule
\textbf{PSNR} & 33.15 & 33.29 & 33.32 & 33.24 & 33.16 \\
\textbf{BPS}  & 0.281  & 0.207   & 0.196   & 0.189  & 0.184  \\
\bottomrule
\end{tabular}
\vspace{-6pt}   %
\end{table}

\emph{Effect of the temporal prediction loss weight $\lambda_{\mathrm{tp}}$.}
Figure~\ref{fig:ablation}(b) sweeps $\lambda_{\mathrm{tp}} \in \{0,\,0.01,\,0.1,\,1,\,10\}$ with $z=256$ on NJCFA. The best tested value $\lambda_{\mathrm{tp}}=0.1$ achieves 37.48 dB at 1.01 BPS, outperforming the no-loss baseline ($\lambda_{\mathrm{tp}}=0$, 37.39 dB at 1.03 BPS) by 0.09 dB while simultaneously reducing bitrate by 0.02 BPS. The improvement is modest in absolute PSNR but reflects a consistent shift of the operating point toward lower BPS. At $\lambda_{\mathrm{tp}}=10$, BPS decreases to 0.93 at the cost of a 0.71 dB PSNR drop relative to $\lambda_{\mathrm{tp}}=0$. This effect is specific to larger latent dimensions ($z=256$); at $z=32$ the impact is negligible, suggesting the temporal prediction loss is most useful when the residual encoder has sufficient capacity to benefit from a well-trained predictor.
\subsection{Real-Time Throughput}
\label{sec:thro_put}
\begin{figure}[t]
\centering
\includegraphics[width=0.9\linewidth]{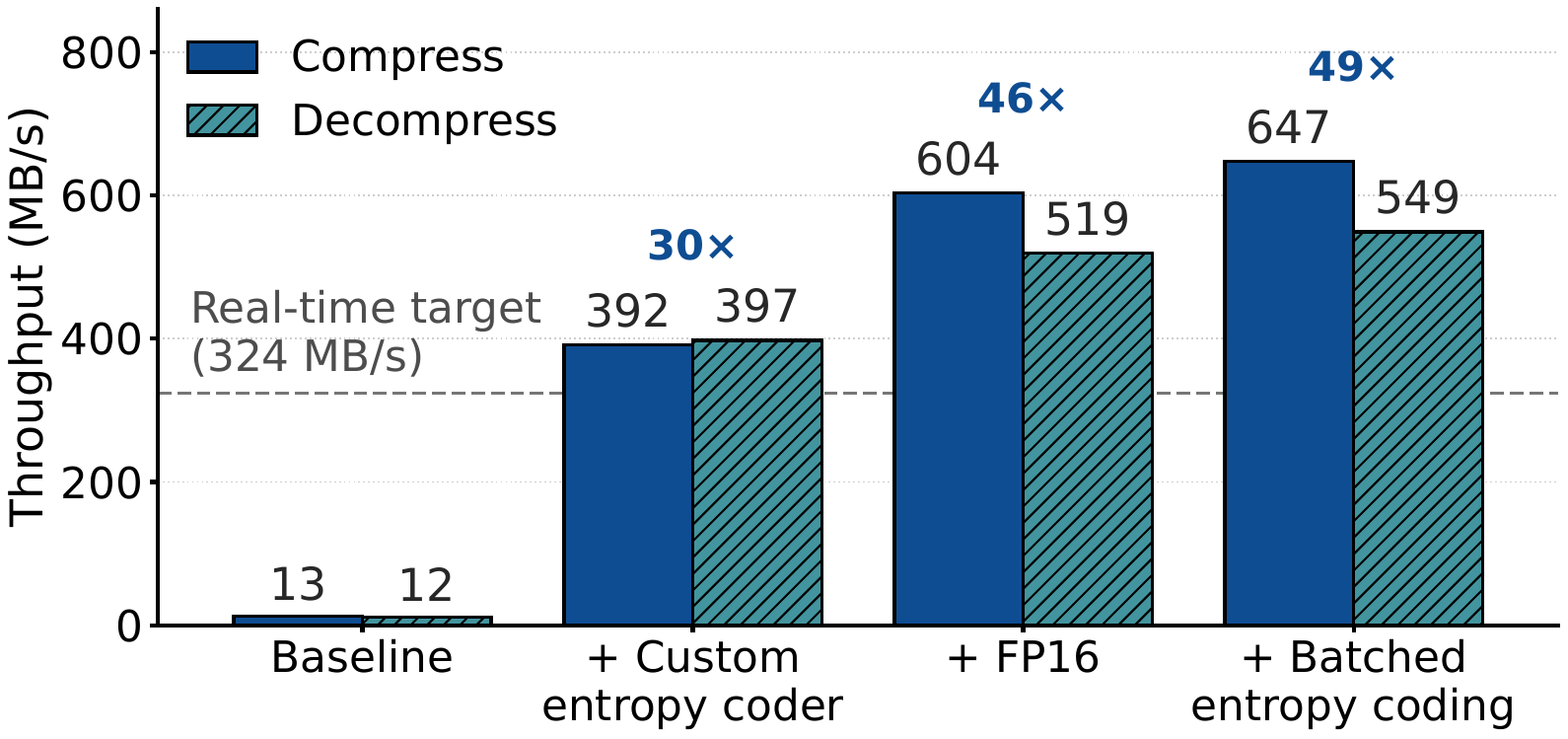}
\vspace{-6pt}
\caption{Cumulative throughput gains from successive optimizations of STeP-C-small model.}
\label{fig:throu_abl}
\vspace{-12pt}
\end{figure}

We first evaluate compression throughput in a simulated streaming setting, where pre-recorded DAS data are supplied to the compression pipeline at controlled rates. This isolates the computational throughput of the compressor from live acquisition and file handling overhead. Figure \ref{fig:throu_abl} summarizes the cumulative throughput gains achieved by each optimization applied to STeP-C-small model. The unoptimized baseline using CompressAI's standard arithmetic coder achieves only 13 MB/s on encoding, far below the 324 MB/s real-time requirement. The custom entropy coder alone raises this to 392 MB/s, a 30× speedup. Adding FP16 mixed-precision inference raises encoding to 604 MB/s. Batching entropy coding across time steps reaches 647 MB/s. The 30$\times$ gain from the custom coder identifies standard entropy coding as the primary bottleneck in the unoptimized pipeline, while FP16 inference and batched coding provide additional throughput gains once this bottleneck is removed.

\subsection{Live Testbed Evaluation}
\label{sec:exp_live}

We next evaluate STeP-C on the PFT dataset under live acquisition. Unlike the simulated streaming evaluation in Section~\ref{sec:thro_put}, this experiment measures the complete pipeline while data are continuously acquired from a DAS interrogator. The model is trained on a 256-channel sub-band and then applied without retraining to one hour of streamed data, comprising 1,111 consecutively acquired files of 3,240 time frames each, sampled at 1 kHz across 81,000 channels. 
Across the full acquisition, STeP-C compresses 1.166 TB of raw DAS data to 13.36 GB, corresponding to an 87.30$\times$ compression ratio and a 98.85\% reduction in file size at 0.367 BPS. Reconstruction on the trained sub-band reaches 33.64 dB PSNR, while full-band reconstruction achieves 31.04 dB. The model therefore remains usable across the full acquisition footprint without channel-specific retraining, though reconstruction quality degrades measurably outside the trained channel range. Including streaming preprocessing, neural compression, parallel entropy coding, and file I/O, the end-to-end live pipeline achieves an average processing throughput of 535 MB/s over the live acquisition, exceeding the 324 MB/s throughput required for 1,000 Hz acquisition across 81,000 channels, with no dropped acquisition frames during the evaluated run.

\begin{table}[t]
\centering
\caption{Downstream detector consistency on DADAS.}
\vspace{-6pt}
\label{tab:downstream}
\small
\begin{tabular}{lc}
\toprule
\textbf{Metric} & \textbf{Value} \\
\midrule
\multicolumn{2}{l}{\textit{Prediction consistency}} \\
Overall top-1 class agreement       & 83.01\% \\
Overall top-5 class agreement       & 98.72\% \\
High-confidence top-1 agreement & \textbf{96.62\%} \\
Probability score correlation        & 0.851 \\
Mean probability deviation  & 0.079 \\
\midrule
\multicolumn{2}{l}{\textit{Detection consistency}} \\
Detection count ratio (recon / org) & 92.13\% \\
Detected class Jaccard similarity   & \textbf{100.00\%} \\
\bottomrule
\end{tabular}%
\vspace{-14pt}   %
\end{table}
\vspace{-6pt}   %
\subsection{Downstream Detector Consistency}
\begin{figure}[t]
\centering
\includegraphics[width=0.8\linewidth]{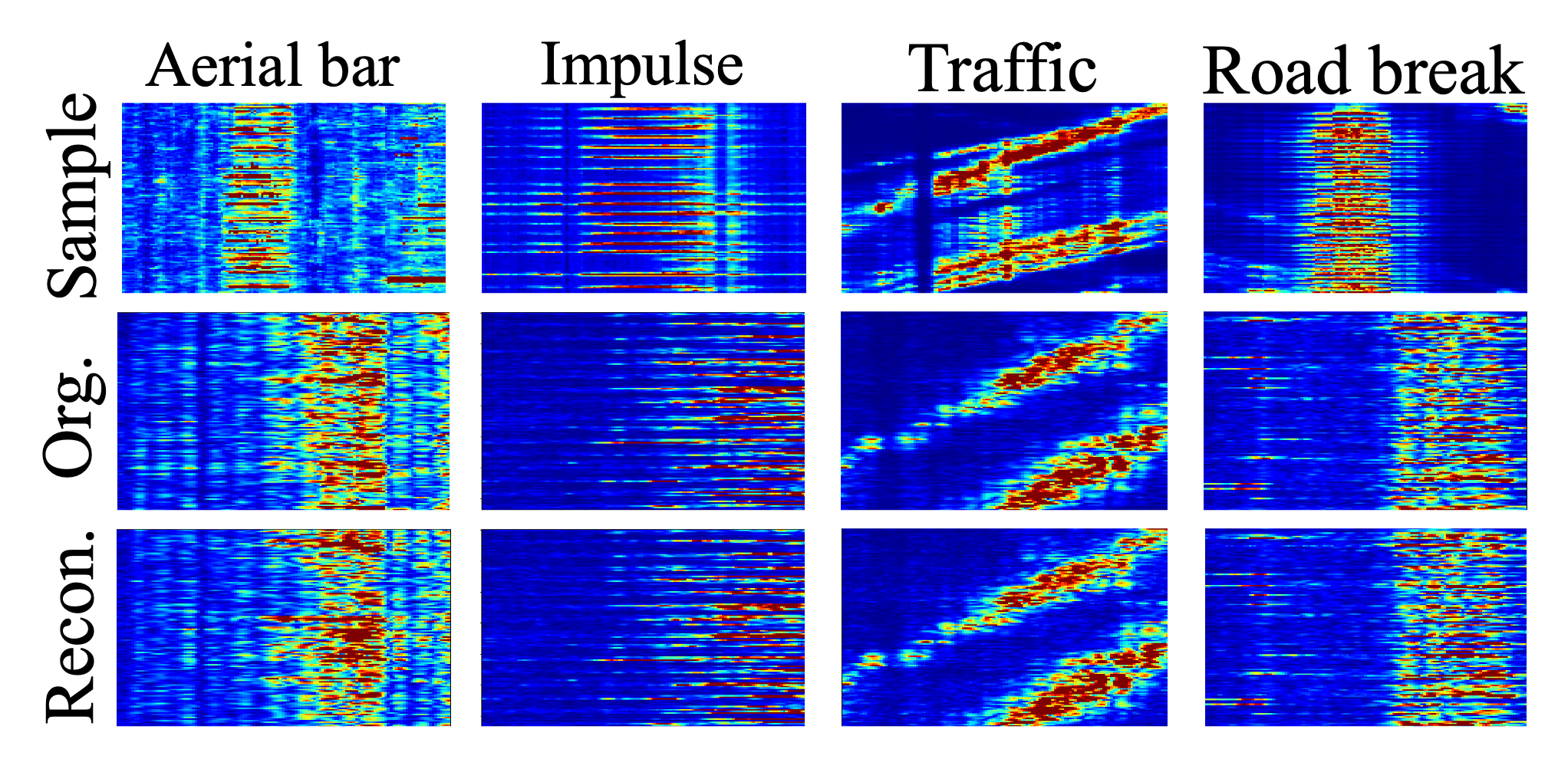}
\vspace{-6pt}
\caption{Qualitative comparison of downstream detections on DADAS. Each column shows an event class; rows show the labeled DAS sample and detector outputs from the original and reconstructed data.}
\label{fig:detection_examples}
\end{figure}

\begin{figure}
    \centering
    \includegraphics[width=0.99\linewidth]{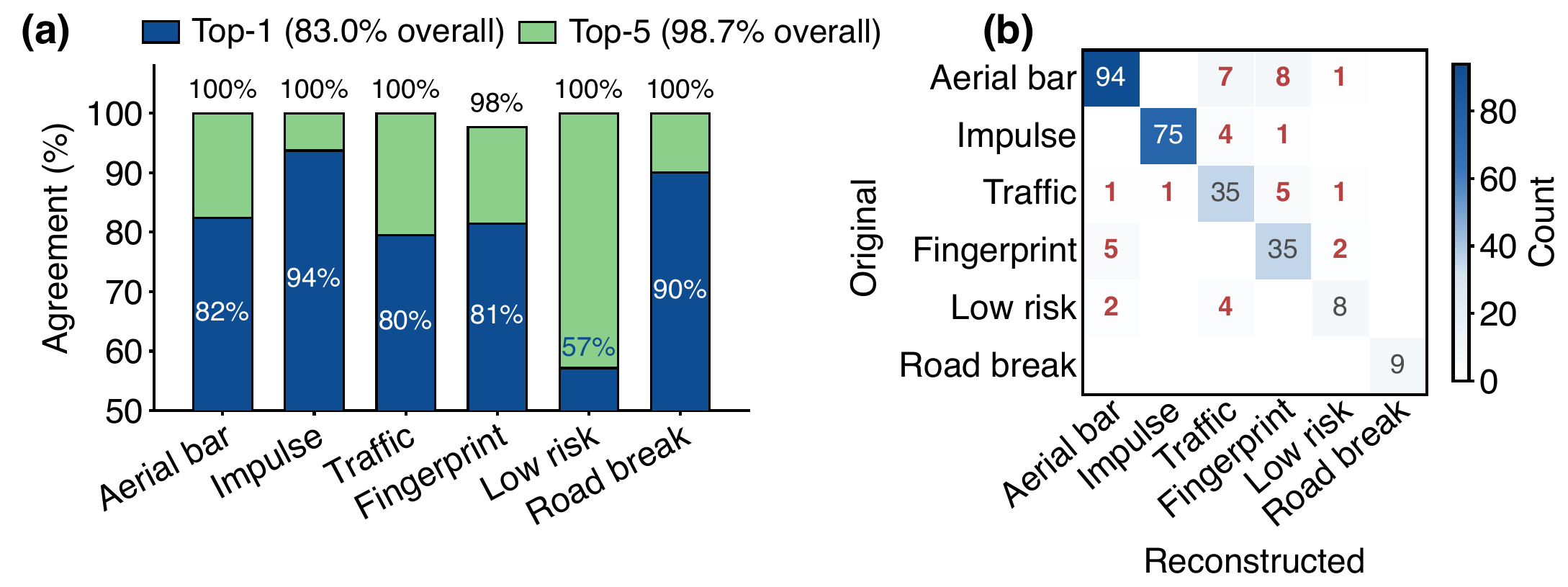}
    \vspace{-6pt}
    \caption{Downstream event detection consistency between original and reconstructed data streams on DADAS. 
    \textbf{(a)} Per-class top-1 (blue) and top-5 (green) agreement for the six most frequent event classes. \textbf{(b)} Class confusion matrix of top-1 predictions between original and reconstructed streams for the same classes.}
    \label{fig:class_agg}
    \vspace{-15pt}
\end{figure}

We assess the impact of compression on the DADAS dataset using a consistency-based protocol in which the same pre-trained detector~\cite{han2024deep} is applied independently to the original and reconstructed data. Before detector inference, both streams undergo the same detector-specific preprocessing. We invert the tanh
normalization, apply the detector's temporal filtering, and aggregate every 30 consecutive samples by averaging their absolute amplitudes. Quantitative results are summarized in Table~\ref{tab:downstream}. Figure~\ref{fig:detection_examples} provides qualitative examples showing that representative event patterns remain visually consistent after reconstruction and lead to the same detector outputs. Across all event windows, the overall top-1 and top-5 class agreements are $83.01\%$ and $98.72\%$, respectively. Restricting the comparison to high-confidence predictions ($p \geq 0.6$) increases top-1 agreement to $96.62\%$, showing substantially higher consistency among confident detector outputs. The corresponding prediction scores remain strongly correlated ($r=0.851$), with a mean probability deviation of $0.079$. Per-class top-1 and top-5 agreements for the six most frequent event classes are shown in Figure \ref{fig:class_agg}(a), while Figure \ref{fig:class_agg}(b) shows the corresponding top-1 confusion matrix. Using the same confidence threshold for event detection, the reconstructed data produce $92.13\%$ as many detections as the original, while the set of detected event classes remains unchanged, yielding a Jaccard similarity of $100\%$. These results indicate that STeP-C largely preserves the pretrained detector’s output behavior, including high-confidence predictions and the set of event categories identified from the original data.


\section{Conclusion}
We presented STeP-C, a neural compression framework for DAS data, and demonstrated its end-to-end deployment under live acquisition. The system achieves an average throughput of 535 MB/s with 87.30\(\times\) compression on continuously streamed interrogator data and 31.04 dB full-band PSNR. Across offline evaluations, STeP-C provides favorable rate-distortion performance in the low-bitrate regime, while temporal prediction consistently reduces the bitrate required for a given reconstruction quality. Downstream evaluation further shows strong consistency between detector outputs on original and reconstructed data, reaching 96.62\% high-confidence class agreement and 100\% Jaccard similarity over detected event classes. These results demonstrate that learned predictive compression can support high-ratio, real-time DAS streaming while retaining both reconstruction quality and downstream utility. Future work will explore spatial representations that generalize more uniformly across fiber locations and extend downstream validation to additional tasks such as event localization and arrival-time picking.

\section*{Acknowledgments}
We acknowledge the use of LLMs to help write data\-processing and
visualization code and to improve the wording of the manuscript.

{
    \small
    \bibliographystyle{ieeenat_fullname}
    \bibliography{main}

\begin{thebibliography}{28}
\providecommand{\natexlab}[1]{#1}
\providecommand{\url}[1]{\texttt{#1}}
\expandafter\ifx\csname urlstyle\endcsname\relax
  \providecommand{\doi}[1]{doi: #1}\else
  \providecommand{\doi}{doi: \begingroup \urlstyle{rm}\Url}\fi

\bibitem[Ajo-Franklin et~al.(2019)Ajo-Franklin, Dou, Lindsey, Monga, Tracy,
  Robertson, Rodriguez~Tribaldos, Ulrich, Freifeld, Daley,
  et~al.]{ajo2019distributed}
Jonathan~B Ajo-Franklin, Shan Dou, Nathaniel~J Lindsey, Inder Monga, Chris
  Tracy, Michelle Robertson, Veronica Rodriguez~Tribaldos, Craig Ulrich, Barry
  Freifeld, Thomas Daley, et~al.
\newblock Distributed acoustic sensing using dark fiber for near-surface
  characterization and broadband seismic event detection.
\newblock \emph{Scientific reports}, 9\penalty0 (1):\penalty0 1328, 2019.

\bibitem[Ballé et~al.(2018)Ballé, Minnen, Singh, Hwang, and
  Johnston]{ball2018variational}
Johannes Ballé, David Minnen, Saurabh Singh, Sung~Jin Hwang, and Nick
  Johnston.
\newblock Variational image compression with a scale hyperprior.
\newblock In \emph{International Conference on Learning Representations}, 2018.

\bibitem[B{\'e}gaint et~al.(2020)B{\'e}gaint, Racap{\'e}, Feltman, and
  Pushparaja]{begaint2020compressai}
Jean B{\'e}gaint, Fabien Racap{\'e}, Simon Feltman, and Akshay Pushparaja.
\newblock {CompressAI}: a {PyTorch} library and evaluation platform for
  end-to-end compression research.
\newblock \emph{arXiv preprint arXiv:2011.03029}, 2020.

\bibitem[Chen et~al.(2025)Chen, Saad, Chen, and Savvaidis]{10833801}
Yangkang Chen, Omar~M. Saad, Yunfeng Chen, and Alexandros Savvaidis.
\newblock Deep learning for seismic data compression in distributed acoustic
  sensing.
\newblock \emph{IEEE Transactions on Geoscience and Remote Sensing},
  63:\penalty0 1--14, 2025.

\bibitem[Cheng et~al.(2019)Cheng, Sun, Takeuchi, and Katto]{cheng2019learning}
Zhengxue Cheng, Heming Sun, Masaru Takeuchi, and Jiro Katto.
\newblock Learning image and video compression through spatial-temporal energy
  compaction.
\newblock In \emph{Proceedings of the IEEE/CVF Conference on Computer Vision
  and Pattern Recognition}, pages 10071--10080, 2019.

\bibitem[Djelouah et~al.(2019)Djelouah, Campos, Schaub-Meyer, and
  Schroers]{9009574}
Abdelaziz Djelouah, Joaquim Campos, Simone Schaub-Meyer, and Christopher
  Schroers.
\newblock Neural inter-frame compression for video coding.
\newblock In \emph{2019 IEEE/CVF International Conference on Computer Vision
  (ICCV)}, pages 6420--6428, 2019.

\bibitem[Dong et~al.(2022)Dong, Popescu, Tribaldos, Byna, Ajo-Franklin, and
  Wu]{DONG2022105181}
Bin Dong, Alex Popescu, Verónica~Rodríguez Tribaldos, Suren Byna, Jonathan
  Ajo-Franklin, and Kesheng Wu.
\newblock Real-time and post-hoc compression for data from distributed acoustic
  sensing.
\newblock \emph{Computers \& Geosciences}, 166:\penalty0 105181, 2022.

\bibitem[Défossez et~al.(2023)Défossez, Copet, Synnaeve, and
  Adi]{defossez2022highfi}
Alexandre Défossez, Jade Copet, Gabriel Synnaeve, and Yossi Adi.
\newblock High fidelity neural audio compression.
\newblock \emph{Transactions on Machine Learning Research}, 2023.

\bibitem[Gemeinhardt et~al.(2026)Gemeinhardt, Sharma, and
  Kastoryano]{gemeinhardt2026quantum}
Hayden Gemeinhardt, Jyotsna Sharma, and Michael Kastoryano.
\newblock Quantum-inspired workflow for processing distributed fiber-optic
  sensor data.
\newblock \emph{Scientific Reports}, 16\penalty0 (1):\penalty0 14972, 2026.

\bibitem[Han et~al.(2024)Han, Huang, Li, Fang, Jiang, and Wang]{han2024deep}
Shaobo Han, Ming-Fang Huang, Tingfeng Li, Jian Fang, Zhuocheng Jiang, and Ting
  Wang.
\newblock Deep learning-based intrusion detection and impulsive event
  classification for distributed acoustic sensing across telecom networks.
\newblock \emph{Journal of Lightwave Technology}, 42\penalty0 (12):\penalty0
  4167--4176, 2024.

\bibitem[Huang et~al.(2021)Huang, Han, Wellbrock, Xia, Narisetty, Salemi, Chen,
  Moore, Ji, Milione, et~al.]{huang2021field}
Ming-Fang Huang, Shaobo Han, Glenn~A Wellbrock, Tiejun~J Xia, Chaitanya
  Narisetty, Milad Salemi, Yuheng Chen, James~M Moore, Philip~N Ji, Giovanni
  Milione, et~al.
\newblock Field trial of cable safety protection and road traffic monitoring
  over operational {5G} transport network with fiber sensing and on-premise
  {AI} technologies.
\newblock In \emph{OptoElectronics and Communications Conference}, pages
  T5A--8. Optica Publishing Group, 2021.

\bibitem[Ip et~al.(2022)Ip, Fang, Li, Wang, Huang, Salemi, and
  Huang]{ip2022distributed}
Ezra Ip, Jian Fang, Yaowen Li, Qiang Wang, Ming-Fang Huang, Milad Salemi, and
  Yue-Kai Huang.
\newblock Distributed fiber sensor network using telecom cables as sensing
  media: technology advancements and applications.
\newblock \emph{Journal of optical communications and networking}, 14\penalty0
  (1):\penalty0 A61--A68, 2022.

\bibitem[Kingma and Ba(2014)]{kingma2014adam}
Diederik~P Kingma and Jimmy Ba.
\newblock Adam: A method for stochastic optimization.
\newblock \emph{arXiv preprint arXiv:1412.6980}, 2014.

\bibitem[Li et~al.(2021)Li, Chen, Huang, Han, and Wang]{li2021vehicle}
Tingfeng Li, Yuheng Chen, Ming-Fang Huang, Shaobo Han, and Ting Wang.
\newblock Vehicle run-off-road event automatic detection by fiber sensing
  technology.
\newblock In \emph{2021 Optical Fiber Communications Conference and Exhibition
  (OFC)}, pages 1--3. IEEE, 2021.

\bibitem[Lin et~al.(2020)Lin, Liu, Li, and Wu]{Lin_2020_CVPR}
Jianping Lin, Dong Liu, Houqiang Li, and Feng Wu.
\newblock M-lvc: Multiple frames prediction for learned video compression.
\newblock In \emph{Proceedings of the IEEE/CVF Conference on Computer Vision
  and Pattern Recognition (CVPR)}, 2020.

\bibitem[Lindsey and Martin(2021)]{lindsey2021fiber}
Nathaniel~J Lindsey and Eileen~R Martin.
\newblock Fiber-optic seismology.
\newblock \emph{Annual review of earth and planetary sciences}, 49\penalty0
  (1):\penalty0 309--336, 2021.

\bibitem[Lindsey et~al.(2019)Lindsey, Dawe, and
  Ajo-Franklin]{lindsey2019illuminating}
Nathaniel~J Lindsey, T~Craig Dawe, and Jonathan~B Ajo-Franklin.
\newblock Illuminating seafloor faults and ocean dynamics with dark fiber
  distributed acoustic sensing.
\newblock \emph{Science}, 366\penalty0 (6469):\penalty0 1103--1107, 2019.

\bibitem[Lu et~al.(2019)Lu, Ouyang, Xu, Zhang, Cai, and Gao]{8953892}
Guo Lu, Wanli Ouyang, Dong Xu, Xiaoyun Zhang, Chunlei Cai, and Zhiyong Gao.
\newblock { DVC: An End-To-End Deep Video Compression Framework }.
\newblock In \emph{2019 IEEE/CVF Conference on Computer Vision and Pattern
  Recognition (CVPR)}, pages 10998--11007, Los Alamitos, CA, USA, 2019. IEEE
  Computer Society.

\bibitem[Ma et~al.(2026)Ma, Meng, and Lin]{ma2026distributed}
Yang Ma, Lingsen Meng, and Yen-Yu Lin.
\newblock Distributed acoustic sensing data compression for seismological
  applications via compressive sensing.
\newblock \emph{Geophysical Journal International}, 246\penalty0 (3):\penalty0
  ggag288, 2026.

\bibitem[Minnen et~al.(2018)Minnen, Ball\'{e}, and Toderici]{3327546.3327736}
David Minnen, Johannes Ball\'{e}, and George Toderici.
\newblock Joint autoregressive and hierarchical priors for learned image
  compression.
\newblock In \emph{Proceedings of the 32nd International Conference on Neural
  Information Processing Systems}, page 10794–10803, Red Hook, NY, USA, 2018.
  Curran Associates Inc.

\bibitem[Ni et~al.(2024)Ni, Denolle, Shi, Lipovsky, Pan, and
  Kutz]{2024JGRH....1..247N}
Yiyu Ni, Marine~A. Denolle, Qibin Shi, Bradley~P. Lipovsky, Shaowu Pan, and
  J.~Nathan Kutz.
\newblock Wavefield reconstruction of distributed acoustic sensing: Lossy
  compression, wavefield separation, and edge computing.
\newblock \emph{Journal of Geophysical Research: Machine Learning and
  Computation}, 1\penalty0 (3):\penalty0 e2024JH000247, 2024.

\bibitem[Seguí et~al.(2026)Seguí, Ugalde, Fichtner, Ventosa, and
  Morros]{10.1093/gji/ggaf397}
Aleix Seguí, Arantza Ugalde, Andreas Fichtner, Sergi Ventosa, and Josep~Ramon
  Morros.
\newblock {DASPack}: controlled data compression for distributed acoustic
  sensing.
\newblock \emph{Geophysical Journal International}, 244\penalty0 (1):\penalty0
  ggaf397, 2026.

\bibitem[Shen et~al.(2024)Shen, Wu, Zhu, Liu, Li, Zheng, Li, Shao, Shum, and
  Lu]{10229168}
Xingliang Shen, Huan Wu, Kun Zhu, Huanhuan Liu, Yujia Li, Hua Zheng, Jialong
  Li, Liyang Shao, Perry~Ping Shum, and Chao Lu.
\newblock Fast and storage-optimized compressed domain vibration detection and
  classification for distributed acoustic sensing.
\newblock \emph{Journal of Lightwave Technology}, 42\penalty0 (1):\penalty0
  493--499, 2024.

\bibitem[Sun et~al.(2025)Sun, Xu, Zhang, Lu, Shi, Zhang, Mu, Qiu, and
  Shao]{sun2025ans}
Jiayao Sun, Deyu Xu, Jingming Zhang, Shanfan Lu, Xiaobing Shi, Peide Zhang, Qi
  Mu, Chaoxian Qiu, and Liyang Shao.
\newblock {ANS-coded} high-ratio data compression for a distributed acoustic
  sensing system.
\newblock \emph{Optics Express}, 33\penalty0 (22):\penalty0 46969--46987, 2025.

\bibitem[Wang et~al.(2024)Wang, Yang, Zeng, Liu, Wang, and Wang]{Wang:24}
Honghui Wang, Xike Yang, Shangkun Zeng, Tong Liu, Xiang Wang, and Yiru Wang.
\newblock Deep learning model for fiber optic distributed acoustic sensing
  seismic data compression based on autoencoder and recurrent neural networks.
\newblock \emph{Appl. Opt.}, 63\penalty0 (20):\penalty0 5303--5315, 2024.

\bibitem[Wang et~al.(2025)Wang, Jiang, Liu, Zhang, Wang, Xu, Zhang, Ding, and
  Liu]{wang2025ultra}
Yixuan Wang, Junfeng Jiang, Kun Liu, Mingjiang Zhang, Shuang Wang, Tianhua Xu,
  Xuezhi Zhang, Zhenyang Ding, and Tiegen Liu.
\newblock Ultra compression-ratio distributed acoustic sensing system via
  combination of envelope extraction and compressed sensing.
\newblock \emph{Optics Express}, 33\penalty0 (5):\penalty0 9854--9864, 2025.

\bibitem[Wellbrock et~al.(2019)Wellbrock, Xia, Huang, Chen, Salemi, Huang, Ji,
  Ip, and Wang]{wellbrock2019first}
Glenn~A Wellbrock, Tiejun~J Xia, Ming-Fang Huang, Yuheng Chen, Milad Salemi,
  Yue-Kai Huang, Philip Ji, Ezra Ip, and Ting Wang.
\newblock First field trial of sensing vehicle speed, density, and road
  conditions by using fiber carrying high speed data.
\newblock In \emph{2019 Optical Fiber Communications Conference and Exhibition
  (OFC)}, pages 1--3. IEEE, 2019.

\bibitem[Zhan(2020)]{zhan2020distributed}
Zhongwen Zhan.
\newblock Distributed acoustic sensing turns fiber-optic cables into sensitive
  seismic antennas.
\newblock \emph{Seismological Research Letters}, 91\penalty0 (1):\penalty0
  1--15, 2020.

\end{thebibliography}
}

\end{document}